\documentclass[runningheads]{llncs}

\usepackage[T1]{fontenc}
\usepackage{graphicx}

\usepackage{amsmath}  
\usepackage{amsfonts}  
\usepackage[colorlinks,
            linkcolor=red,       
            anchorcolor=yellow,    
            citecolor=blue,      
            urlcolor=green,
            ]{hyperref}
\usepackage{float}
\usepackage{multirow}
\usepackage{booktabs}
\usepackage{svg} 

\usepackage{perpage}
\MakePerPage{footnote}

\usepackage{caption}  
\usepackage{subcaption}

\usepackage{marvosym}

\begin{document}

\title{External Prompt Features Enhanced Parameter-efficient Fine-tuning for Salient Object Detection}
\titlerunning{ExPert for SOD}

\author{Wen Liang\inst{1} \and  
Peipei Ran\inst{2} \and  
Mengchao Bai\inst{2} \and 
Xiao Liu\inst{2} \and  
P. Bilha Githinji \inst{1} \and   
Wei Zhao \inst{2} \and  
Peiwu Qin\inst{1}\textsuperscript{\Letter}}  

\authorrunning{W. Liang et al.}
%
\institute{Tsinghua Shenzhen International Graduate School, Tsinghua University, Shenzhen, China \and
Central Media Technology Institute, Huawei, Shenzhen, China
}

\maketitle              

\begin{abstract}
Salient object detection (SOD) aims at finding the most salient objects in images and outputs pixel-level binary masks. Transformer-based methods achieve promising performance due to their global semantic understanding, crucial for identifying salient objects. However, these models tend to be large and require numerous training parameters. To better harness the potential of transformers for SOD, we propose a novel parameter-efficient fine-tuning method aimed at reducing the number of training parameters while enhancing the salient object detection capability. Our model, termed \textbf{EX}ternal \textbf{P}rompt features \textbf{E}nhanced adapte\textbf{R} \textbf{T}uning (ExPert), features an encoder-decoder structure with adapters and injectors interspersed between the layers of a frozen transformer encoder. The adapter modules adapt the pre-trained backbone to SOD while the injector modules incorporate external prompt features to enhance the awareness of salient objects. Comprehensive experiments demonstrate the superiority of our method. Surpassing former state-of-the-art (SOTA) models across five SOD datasets, ExPert achieves 0.215 mean absolute error (MAE) in the ECSSD dataset with 80.2M trained parameters, 21\% better than SelfReformer\cite{yun2022selfreformer} and 47\% better than EGNet\cite{zhao2019egnet}. 

\keywords{Salient object detection  \and Segmentation \and Adapter tuning\and Prompt tuning \and Vision language model.}
\end{abstract}

\section{Introduction}

\subsection{Motivation}

Salient object detection (SOD) is a widely studied task in computer vision that outputs a binary mask of the visually salient objects in an image. The detection of salient objects can benefit various computer vision tasks, such as semantic segmentation, instance segmentation, and object detection. In recent years, convolutional neural network (CNN) based models and transformer-based models have shown promising performances for SOD. However, although transformer-based models~\cite{yun2022selfreformer,ren2021unifying_GLSTR,liu2021visual_salient} generally outperform their CNN counterparts, they are more computationally expensive due to their typically large number of parameters that are essential for achieving superior performance.

The encoder-decoder framework is widely used for salient object detection, which is defined as a binary semantic segmentation. Firstly, a vision encoder is initialized with pre-trained model weights from classification or segmentation models. The next step is to fine-tune the encoder and decoder on salient object detection datasets to extend the model to the SOD task. The predicted salient masks are generated by the specific decoder with the extracted features. Beyond fine-tuning, training the pre-trained backbone along with other new sophisticated modules can gain better performance. However, it necessitates an even larger number of trained parameters. 

To fine-tune pre-trained transformer models efficiently with fewer parameters, we leverage adapter tuning~\cite{chen2022adaptformer} that selectively fine-tunes certain side connections within frozen transformer blocks, facilitating transferability to downstream tasks. However, only manipulating features of the frozen backbone does not effectively tackle the salient object detection task. In \cite{jia2022visual_VPT}, some learnable prompt vectors are added to the transformer layers to fine-tune large transformer models for specific tasks. Inspired by the effectiveness of visual prompt tuning, we assume that features from external backbones can be employed as prompt features. The injection of suitable external prompt features can enhance the performance of SOD models in addition to adapter tuning.

\subsection{Methods Overview}

We propose \textbf{EX}ternal \textbf{P}rompt features \textbf{E}nhanced adapte\textbf{R} \textbf{T}uning (ExPert) model to parameter-efficiently tune pre-trained transformer backbones for salient object detection. ExPert is a backbone-agnostic model and can be extended to any transformer-based pre-trained backbones. Inspired by~\cite{chen2022adaptformer,liu2023explicit}, ExPert uses the block-level\footnote{In~\cite{chen2022adaptformer}, the adapter is a side connection of the feed-forward function inside the transformer block which is denoted as "FFT-level". In~\cite{liu2023explicit} and our ExPert, the adapter is a side connection between transformer blocks and is denoted as "block-level".} adapter module to tune the transformer backbone between each block unit. We denote the adapter of ExPert as E-adapter.

We also design a block-level injector module E-injector to receive external prompt features and inject them into the backbone so as to enhance salient features. The encoder backbone is frozen during training while the E-adapters, the E-injectors and the decoder are trained. The vision features from DINO~\cite{caron2021emerging}, ViT~\cite{dosovitskiy2020image_ViT} and BLIP~\cite{li2022blip} are chosen to verify the compatibility of our E-injector. 

Moreover, we hypothesize that the captions of images are highly related to the salient elements. Based on this premise, ExPert interacts BLIP's visual features and text embeddings of corresponding captions using cross attention. The best result was achieved by injecting the interacted features combined with ViT's features into the backbone. Comprehensive experiments show that ExPert surpasses CNN-based SOTA models largely and performs better than previous transformer-based models.

\subsection{Contributions}

Our main contributions lie in three aspects: 
\begin{itemize}
  \item[$\bullet$] We propose the ExPert model to parameter-efficiently fine-tune pre-trained transformer backbones for salient object detection. ExPert is backbone-agnostic and can use different transformer-based backbones. Comprehensive experiments demonstrate the superiority of ExPert.
  \item[$\bullet$] We design the block-level E-adapter to parameter-efficiently adapt the pre-trained transformer backbones to salient object detection. The size of the trained parameters of ExPert is only 80.2M.
  \item[$\bullet$] Our E-injector can receive different external prompt features and inject them to guide the backbone to extract salient features. Experiments demonstrate that the injection of features that contain rich semantic information largely boosts the performance.
\end{itemize}

\section{Related Work}

\subsection{Salient Object Detection}
CNN-based models are proficient at extracting local details. EGNet \cite{zhao2019egnet} focuses on the complementarity between edges and the content of salient objects by extracting edge information. U2Net \cite{qin2020u2_u2net} proposes a nested U-shape convolutional network to handle inputs with flexible sizes without any pre-training. Although requiring more computing costs, transformer-based models surpass CNN-based models in SOD because transformer models can grasp the long-range semantic context of input images. SelfReformer~\cite{yun2022selfreformer} adopts a global branch to refine the local context branch with a multi-stage transformer backbone to achieve better long-range information extraction. EVP \cite{liu2023explicit} fine-tunes SegFormer~\cite{xie2021segformer} with patch embedding prompts and Fourier transformation prompts to better differentiate objects.

\subsection{Adapter Tuning and Visual Prompt Tuning}
Adapter tuning is a method to fine-tune pre-trained models which was first proposed in~\cite{rebuffi2017learning_adapter_origin} as a trainable side connection branch for parameter-efficient tuning. Later Houlsby et al.~\cite{houlsby2019parameter_adapter} used adapter tuning to parameter-efficiently train transformer-based language models. AdapterFormer~\cite{chen2022adaptformer} applies adapter tuning to Vision Transformer and achieves promising performance in multi-label classification. EVP~\cite{liu2023explicit} demonstrates that adapter tuning can effectively transfer pre-trained transformer-based models to downstream tasks such as salient object detection, camouflaged object detection, and other binary segmentation tasks.

Prompt~\cite{liu2023pre_prompt} is originally used in natural language processing (NLP) to instruct pre-trained language models to understand and shift to new tasks. Prompt tuning has also developed rapidly in the computer vision (CV) domain. An input-agnostic visual perturbation prompt is learned and fed to a model together with input images to repurpose pre-trained models to downstream tasks in~\cite{bahng2022exploring_VisualPrompt}. Some learnable parameters are injected into the transformer's input space to efficiently fine-tune large-scale transformer models in~\cite{jia2022visual_VPT}. ViT-Adapter~\cite{chen2022vision_ViT_adapter} uses side branches to inject spatial priors into ViT to fine-tune the model for detection and segmentation tasks. These works all show that the injection of prompt information into the original backbone can guide pre-trained models for versatile downstream tasks. 

\section{Methods}

\begin{figure*}[htbp] 
\centering 
\includegraphics[width=\textwidth]{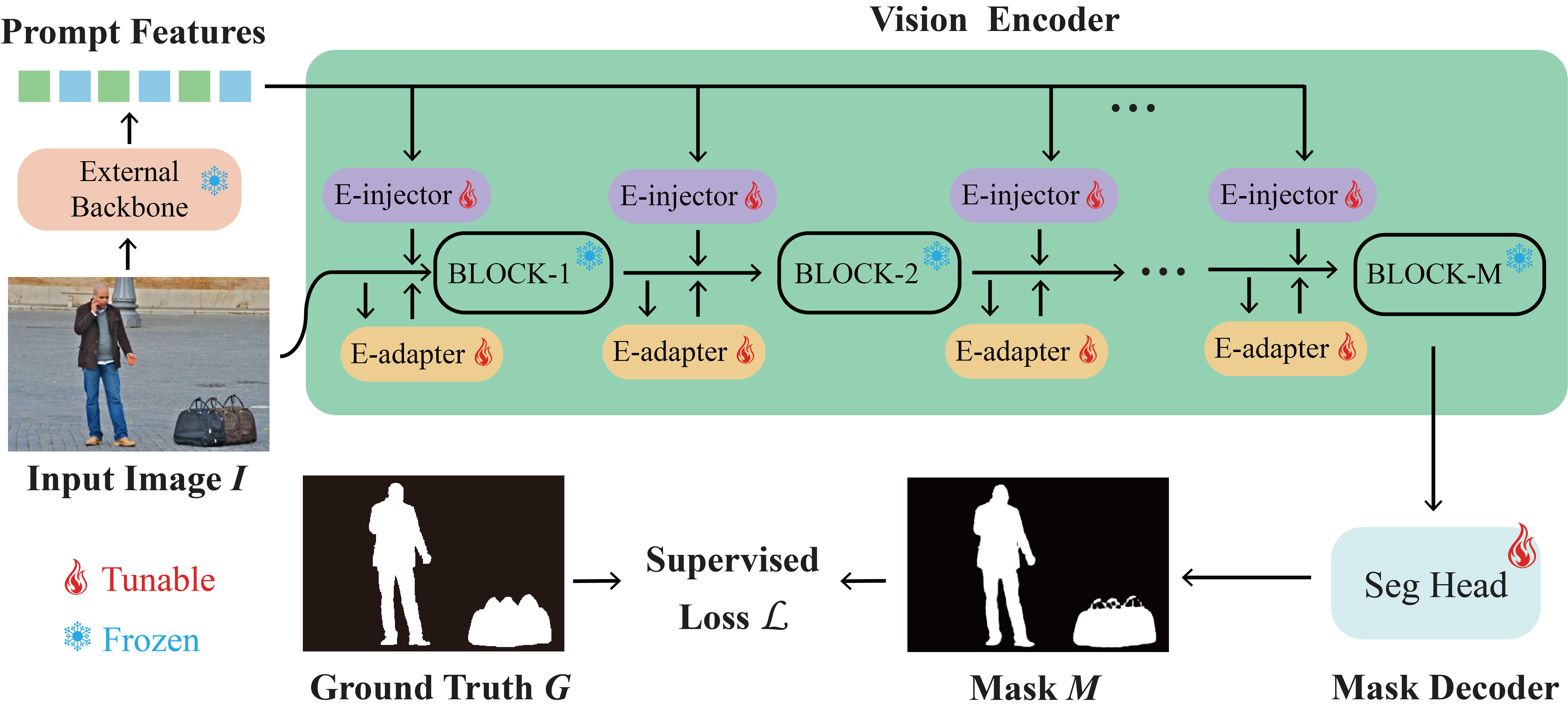}
\caption{The overall architecture of ExPert. During training, the vision encoder is frozen; only the E-adapters, E-injectors and the decoder are trained. } 
\label{Overview} 
\end{figure*}

\subsection{Overview}

Salient object detection is an important task in the computer vision field, which detects the most salient objects in an RGB image and outputs binary masks of these objects. Let $I \in \mathbb{R}^{H\times W\times 3}$ denote the input image and $G \in \mathbb{R}^{H\times W\times 1}$ the corresponding ground truth binary mask. The output binary mask of the model is $M \in \mathbb{R}^{H\times W\times 1}$. Suppose the SOD model is $\mathcal{F}$ and its parameters are $\theta$, then the mask is calculated as $\mathcal{F}(I,\theta)$. The loss function $\mathcal{L}$ in ExPert is a combination of binary cross entropy (BCE) loss and intersection over union (IoU) loss~\cite{yu2016unitbox_iou}. The training target is to minimize $\mathcal{L}(M, G)$ between $M$ and $G$.

We propose an encoder-decoder model denominated as \textbf{EX}ternal \textbf{P}rompt features \textbf{E}nhanced adapte\textbf{R} \textbf{T}uning (ExPert) with block-level E-adapter and E-injector. The architecture of ExPert is shown in Fig.~\textcolor{red}{\ref{Overview}} and entails a vision encoder, a mask decoder, some E-adapter modules between the transformer blocks of the vision encoder and some shared E-injectors for each feature scale. 

Since SOD is defined as a segmentation task that is similar to semantic segmentation, pre-trained transformer backbones for segmentation or classification are preferable. A multi-scale encoder of SegFormer~\cite{xie2021segformer} and its decoder are chosen as the backbone and the segmentation head of ExPert. E-adapter is a lightweight side connection module that helps to transfer the pre-trained transformer backbone to salient object detection. In addition, E-injector is a lightweight side connection module that projects vision features from other backbones as guiding prompts and injects these prompts into the encoder. The detailed structures of E-adapter and E-injector are illustrated in Fig.~\ref{Adapter Fig}. 

ExPert is trained in an end-to-end manner with image-mask pairs. During training, the vision encoder is frozen while the E-adapters, E-injectors and decoder are trained. Experiments demonstrate that the combined prompts of BLIP and ViT achieve the best result. As a backbone-agnostic model, ExPert can  switch between different transformer backbones with simple modification\footnote{More details can be found in the supplementary materials.} while the decoder needs to be specified according to different backbones.

\subsection{Encoder and Decoder} \label{encoder}

\begin{figure}[htbp]
    \centering
        \includegraphics[width=350pt]{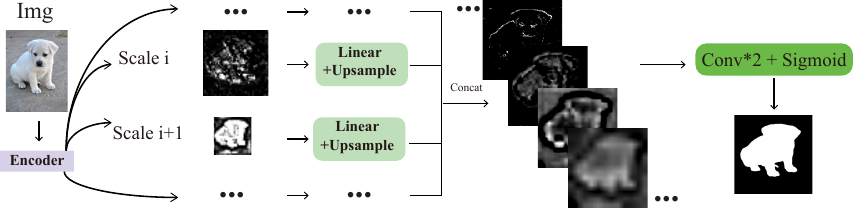} 
    \caption{The decoder of ExPert for multi-scale features. The illustration of feature images is visualized by choosing a random slice of the channel dimension. ExPert's final mask is generated by resizing this mask to the original size.}
    \label{SegFormer Decoder}
\end{figure}

Transformer backbones can be classified into two types according to whether the scale of features changes. One is the single-scale backbone and the other is the multi-scale backbone. The features of the single-scale backbone keep the same size during the forward propagation while the multi-scale backbone's features change size. For salient object detection, former research~\cite{zhao2019pyramid_PFAN,qin2020u2_u2net,yun2022selfreformer} emphasized the importance of multi-scale feature fusion to get finer segmentation masks. Since multi-scale features are crucial to dense prediction for finer details, multi-scale backbones have an advantage over single-scale backbones in segmentation tasks. Therefore we choose the MiT-B4 version of SegFormer~\cite{xie2021segformer} as our multi-scale backbone. The encoder of SegFormer has 4 stages of feature extraction and each stage has a different scale of features. The decoder is kept intact as the original one as shown in Fig.~\ref{SegFormer Decoder}. 

For each stage of feature extraction, denote the output feature of the $i_{th}$ stage as $F^{out}_i$. A linear layer $L_i$ projects $F^{out}_i$ to align the channel dimension for the following concatenation, followed by a bilinear upsampling $\mathcal{U}()$ to align the spatial scale of features. Then all aligned features from different stages are concatenated by $Cat()$ to get $F_c$. Finally, two convolution layers, $C_{fuse}$, $C_{pred}$ and the sigmoid function $\mathcal{S}()$ are applied for the mask generation. The mask $M$ is generated as Eq.~\ref{SegFormer Mask Cal}.

\begin{align}
    F_c &= Cat(\sum^{I}_{i=1} \mathcal{U}(L_i(F^{out}_i))) \\
    M &= \mathcal{S}(C_{pred}(C_{fuse}(F_c)))
\label{SegFormer Mask Cal} 
\end{align}

\subsection{E-adapter} \label{ExPert Adapter Chapter}

\begin{figure}[htbp]
    \centering    
        \includegraphics[width = 300pt]{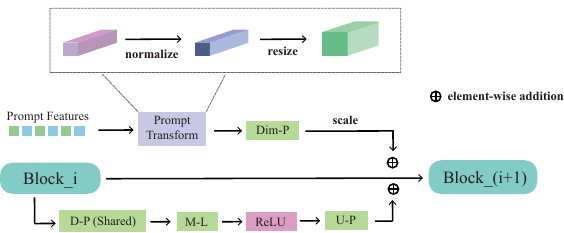}
    \caption{The detailed structure of E-adapter and E-injector. D-P is the down projection layer, M-L is the median linear layer, U-P is the up projection layer and Dim-P is the dimension projection layer.}
    \label{Adapter Fig}
\end{figure}

To fine-tune the transformer backbone in a parameter-efficient manner, we employ the adapter tuning method, which reduces the dimension of features through a bottleneck design, thereby diminishing the number of trained parameters. As shown in the bottom branch of Fig.~\ref{Adapter Fig}, we propose the E-adapter, consisting of a down-projection(D-P) layer with parameters $P^{d}_e$, a median linear(M-L) layer $L^{m}_e$ and an up-projection(U-P) layer with parameters $P^{u}_e$. It is noteworthy that the D-P layers of the E-adapter are shared for features of the same spatial scale in order to further diminish the number of trained parameters. The M-L and U-P layers are independent for each E-adapter. The additional low-dimension M-L layers serve to increase the variability of the E-adapter. Define the $i_{th}$ block forward function as $\mathcal{B}_i()$, the $(i+1)_{th}$ block's feature $F_{i+1}$ after the E-adapter is computed as Eq.~\ref{E-adapter output}.\footnote{The $P \cdot F$ in Eq.~\ref{E-adapter feature} represents the linear projection to features F with parameters P.}

\begin{align} 
    F^{ad}_i &= P^{u}_e \cdot ReLU(L^{m}_{e}(P^{d}_{e} \cdot F_i))  \label{E-adapter feature}  \\
     F_{i+1}  &= \mathcal{B}_i(F_i + F^{ad}_i)  \label{E-adapter output}
\end{align}

\subsection{E-injector} \label{ExPert Injector Chapter}

Previous works~\cite{chen2022vision_ViT_adapter,liu2023explicit} show that side connection modules like adapters can introduce extra information to boost the model's performance in object detection and segmentation tasks. We design the E-injector to inject external features from other backbones of the same input images as guiding prompts into the encoder for salient object detection. The E-injector's structure is depicted in the top branch of Fig.~\ref{Adapter Fig}. 

If the number of injected prompt features is $J$, the $j_{th}$ E-injector is composed of a prompt transformation ${Trans}_j()$ and a dimension projection (Dim-P) layer with parameters $P^{dim}_j$.  Since visual prompts might vary in size and shape, a feature transformation ${Trans}_j()$ fits the prompt feature  $F'_j$ to the $i_{th}$ layer's feature $F_i$ of the frozen backbone. ${Trans}_j()$ is composed of a normalization and a resize operation. The E-injector can receive different transformer features $F'_j$ including the features from DINO, ViT and BLIP's vision encoder. The output of the E-injector $F^{inj}_j$ is generated as Eq.~\ref{ExPert injector output}. To better adjust the injector features to the backbone, a learnable scaling vector $\alpha_j$ is used to weight them. The $(i+1)_{th}$ block's feature $F_{i+1}$ is computed as Eq.~\ref{ExPert adapter}.

\begin{align}
    F^{inj}_j &= P^{dim}_j  \cdot {Trans}_j(F'_j)    \label{ExPert injector output}  \\
      F_{i+1}  &=  \mathcal{B}_i(F_i +  F^{ad}_i + \sum_{j=1}^{J}F^{inj}_j \times \alpha_j )      \label{ExPert adapter} 
\end{align}

Finding suitable prompt features is crucial to the quality of salient masks for E-injector. DINO~\cite{caron2021emerging} is a self-supervised model trained without labels that exhibits an obvious tendency to focus on objects in an image. Observing the attention maps of DINO reveal coarse masks that are nearly similar to the masks of the salient objects, we assume this kind of object-aware features can aid pre-trained models in locating objects. Besides, considering that our multi-scale backbone has fewer layers of high resolution features compared to single-scale backbones such as ViT, our backbone's overall perception of an image might be complemented by the features of ViT's last layer. ExPert uses the features of ViT/B-16 as the auxiliary features for better global perception of images.   

While piling up the features might strengthen the visual details, it is still hard to guide the model to recognize the notion of saliency. It is noteworthy that the caption naturally contains the descriptions of the salient objects in an image which are beneficial to SOD. Therefore, we envision that the caption of an image is highly related to the salient objects and contains the salient information. Some Vision Language Models (VLM) like BLIP~\cite{li2022blip} are trained on large caption datasets.\footnote{CLIP~\cite{radford2021learning_clip} is a well-known VLM model trained with millions of image-text pairs. However the text of CLIP is a simple sentence with the class name which is not the caption of the whole image. The resolution of CLIP's training images is 224*224, the feature is 7*7 with the patch size of 32 which is too small to upsample. Therefore ExPert does not consider CLIP's features.} The features from BLIP's vision encoder trained with image-text labels, which contain rich semantic information of an image, are injected as semantic-enhanced features by E-injector.

To fully explore the rich semantic information in BLIP, we interact BLIP's vision features and BLIP's captions for better focus on salient objects. Although there are no captions or other text information in SOD datasets, we can generate them by the inference results of the BLIP model. For an image $I$, the corresponding caption is generated by BLIP using beam search. The caption is tokenized and embedded by BLIP's text encoder to get the text embedding $T$. The interacted feature is acquired via the cross-attention between the last layer vision feature $V_b$ of BLIP and $T$. In each cross attention layer, $V_b$ is projected as the query and $T$ is projected as the key and value by linear projections. Experiments show that one cross-attention layer is enough for the interaction\footnote{ More details can be found in the supplementary file. }. Since BLIP is trained with image-caption pairs, the alignment of image and text features is not of pixel scale but rather of patch scale. As auxiliary prompt features, too many cross attentions might lead to unexpected noises that harm the final performance.

Since the features from ViT and similar models can refine the mask in detail and the features from BLIP can inject semantic information into ExPert, we combine these two features together for E-injector. The final version of ExPert uses the features from ViT's last layer and the interacted features by cross-attention from BLIP. Experiments show that the combination prompt version achieves the best performance. Unless specified, ExPert represents our best version.

\section{Experiments}

\subsection{Experiment Setup}

Implemented with PyTorch, ExPert is trained on the DUTS \cite{wang2017learning-DUTS} dataset with a batch size of eight using two V100 16G GPUs. The encoder and the decoder are initialized with the publicly released pre-trained weights of SegFormer and the other parameters are initialized randomly. We used the AdamW optimizer and the learning rate is set to 2e-4 with a weight decay of zero.

We used ECSSD \cite{yan2013hierarchical-ECSSD}, DUT-OMRON \cite{yang2013saliency-DUT-OMRON}, HKU-IS \cite{li2015visual-HKU-IS} and PASCAL-S \cite{li2014secrets-PASCAL-S} as the evaluation datasets. Four metrics are adopted for our model evaluation: the mean absolute error (MAE), the F-measure $F_\beta$ \cite{achanta2009frequency_fmeasure}, the maximum E-measure \cite{fan2018enhanced_emeasure} and the S-measure \cite{fan2017structure_smeasure}. More details on implementation, datasets and metrics can be found in the supplement file.

\subsection{Comparison with SOTA Models}

\subsubsection{Quantitative Comparison}

\begin{table}[h]
    \centering
    \begin{subtable}[t]{1.0\linewidth}
        \centering
        \begin{tabular}{l|rrrr|rrrr|rrrr}
            \toprule
            \multirow{2}*{Methods} & \multicolumn{4}{c}{DUTS-TE} & \multicolumn{4}{c}{DUT-OMRON} & \multicolumn{4}{c}{ECSSD} \\
            \cmidrule{2-13}
                   & MAE$\mathbf{\downarrow}$ & FM$\mathbf{\uparrow}$ & EM$\mathbf{\uparrow}$ & SM$\mathbf{\uparrow}$ 
                   & MAE$\mathbf{\downarrow}$ & FM$\mathbf{\uparrow}$ & EM$\mathbf{\uparrow}$ & SM$\mathbf{\uparrow}$ 
                   & MAE$\mathbf{\downarrow}$ & FM$\mathbf{\uparrow}$ & EM$\mathbf{\uparrow}$ & SM$\mathbf{\uparrow}$ \\
            \midrule
           
            EGNet        & .0431                       & .8507                       & .9148                       & .8775                      
                         & .0564                       & .7686                       & .8640                       & .8345                       
                         & .0405                       & .9293                       & .9494                       & .9192                      \\ 
            U2Net        & .0443                       & .8477                       & .9102                       & .8737                      
                         & .0544                       & .7930                       & .8794                       & .8466                       
                         & .0330                       & .9408                       & .9572                       & .9276                      \\
            EVP          & .0297                       & .9033                       & .9521                       & .9016                      
                         & .0485                       & .8195                       & .9047                       & .8529                       
                         & .0303                       & .9475                       & .9636                       & .9335                      \\
            SR           & .0266                       & .9016                       & .9514                       & .9110                      
                         & .0433                       & .8058                       & .8899                       & .8603                       
                         & .0273                       & .9480                       & .9651                       & .9356                    \\
            ExPert        & \textbf{.0231}              & \textbf{.9158}              & \textbf{.9594}              & \textbf{.9179}                      
                         & \textbf{.0429}              & \textbf{.8399}              & \textbf{.9101}              & \textbf{.8711}                       
                         & \textbf{.0215}              & \textbf{.9550}              & \textbf{.9707}              & \textbf{.9422}                      \\ 
            \bottomrule
        \end{tabular}
        \caption{}
    \end{subtable}

    \begin{subtable}[t]{1.0\linewidth}
        \centering
        \begin{tabular}{l|rrrr|rrrr|r}
            \toprule
            \multirow{2}*{Methods} &  \multicolumn{4}{c}{HKU-IS} & \multicolumn{4}{c}{PASCAL-S}  & \multicolumn{1}{c}{\quad\quad\quad TP\quad}\\ 
            \cmidrule{2-10}
                   & MAE$\mathbf{\downarrow}$ & FM$\mathbf{\uparrow}$ & EM$\mathbf{\uparrow}$ & SM$\mathbf{\uparrow}$
                   & MAE$\mathbf{\downarrow}$ & FM$\mathbf{\uparrow}$ & EM$\mathbf{\uparrow}$ & SM$\mathbf{\uparrow}$ 
                   & Size \\
            \midrule
           
            EGNet        & .0345                       & .9160                       & .9520                       & .9098                      
                         & .0821                       & .8166                       & .8673                       & .8469                        
                         & 412.3M\\ 
            U2Net        & .0312                       & .9238                       & .9539                       & .9160                      
                         & .0817                       & .8097                       & .8609                       & .8414                        
                         & 168.1M\\
            EVP          & .0253                       & .9426                       & .9694                       & .9294                      
                         & .0674                       & .8486                       & .8930                       & .8701                        
                         & 14.1M\\
            SR           & .0241                       & .9406                       & .9689                       & .9309                 
                         & .0600                       & .8513                       & .8978                       & .8807                        
                         & 349.7M\\
            ExPert        & \textbf{.0198}              & \textbf{.9498}              & \textbf{.9747}              & \textbf{.9375}                      
                         & \textbf{.0538}              & \textbf{.8670}              & \textbf{.9099}              & \textbf{.8932}               
                         & 80.2M\\ 
            \bottomrule
        \end{tabular}
        \caption{}
    \end{subtable} 
    
    \caption{The quantitative metrics of our best version of ExPert(ViT \& BLIP+ Injection) and four SOTA models. Best results are in \textbf{bold}. SR is the abbreviation of SelfReformer. EVP and SR are two transformer-based SOTA models while EGNet and U2Net are two CNN-based SOTA models. The column of trained parameters (TP) shows the size of trained parameters of each model.}
    \label{SOTA Compare}
\end{table}

We compare our model's salient masks with four representative state-of-the art models on five salient object detection datasets: DUTS-TR, DUT-OMRON, ECSSD, HKU-IS and PASCAL-S. Two CNN SOTA models EGNet \cite{zhao2019egnet} and U2Net \cite{qin2020u2_u2net} are considered together with two transformer SOTA models SelfReformer \cite{yun2022selfreformer} and EVP \cite{liu2023explicit}. The metrics are calculated under the same condition using the prediction masks of different models\footnote{We use the public codes of \href{https://github.com/zyjwuyan/SOD_Evaluation_Metrics}{SOD\_Evaluation\_Metrics} to compute the metrics.}. The prediction masks are all provided by the official release\footnote{Considering that EVP's official mask is 352*352 which is not the original size, we resize the prediction map of EVP to the size of ground truth and then compute the metrics.}. 

\begin{figure}
    \centering
    \includegraphics[width=350pt]{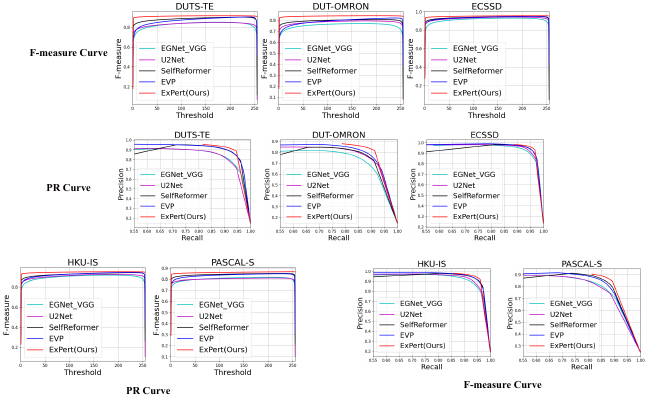}
    \caption{The F-measure curves and the precision-recall (PR) curves of ExPert and four SOTA models on five datasets.}
    \label{Quantitative}
\end{figure}

The results are shown in Table.~\textcolor{red}{\ref{SOTA Compare}}. Our method achieves the best performance across all the SOTA models on all five datasets, which demonstrates the effectiveness of ExPert and the potential of the transformer backbone on salient object detection. For the MAE metric, ExPert surpasses the second best SelfReformer by 0.0058 (around 21\% improvement) in the ECSSD dataset and surpasses EGNet by around 47\%. The superiority on all other three metrics demonstrates that ExPert has stronger competence to segment the salient objects in an image. Regarding the size of trained parameters, ExPert is more parameter efficient than all SOTA models except EVP whose trained parameters are also under 100M as ExPert but with a smaller size. Therefore, ExPert realizes a good trade-off between performance and the size of trained parameters. Fig.~\ref{Quantitative} shows the F-measure curves and the precision-recall curves of ExPert and 4 SOTA models on five datasets. It is observable  in these curves that our model consistently outperforms all other models.

To ensure sufficient comparisons, ExPert is also compared to other latest SOD models\footnote{Due to the absence of codes or salient maps of some models, we directly use the metrics results in the published paper. The results of M3Net\cite{yuan2023m_M3Net} are calculated using the official salient maps of the M3Net SwinB version.} including M3Net\cite{yuan2023m_M3Net}, DSRNet\cite{song2023salient_DSRNet}, TCRNet\cite{zhang2022tcrnet}, BBRF model\cite{ma2023boosting_BBRF}, IMSFNet\cite{xia2023imsfnet} and CTD-L\cite{li2023rethinking_CTD}. As shown in Table.~\ref{More SOTA compare}, ExPert performs better in three metrics than these latest SOD models in the DUT-OMRON dataset, which demonstrates the superiority of ExPert.

\begin{table*}
    \centering
    \resizebox*{1.0\columnwidth}{!}{
        \begin{tabular}{lrrrrrrr}
            \toprule
            Metrics  & ExPert & M3Net\cite{yuan2023m_M3Net} & DSRNet\cite{song2023salient_DSRNet} & TCRNet\cite{zhang2022tcrnet} & BBRF\cite{ma2023boosting_BBRF} & IMSFNet\cite{xia2023imsfnet} & CTD-L\cite{li2023rethinking_CTD}  \\
            \midrule
            MAE$\mathbf{\downarrow}$  & \textbf{.042}  & .045 & .051  & .054 & \textbf{.042} & .053  & .049 \\
            max-FM$\mathbf{\uparrow}$   & \textbf{.839}  & .832 & .810  & 0.791 & .814 & .760  & .789 \\
            max-EM$\mathbf{\uparrow}$   & \textbf{.910}  & .902 & / & / & .887 & .777  & .881 \\
            SM$\mathbf{\uparrow}$   & .871  & \textbf{.872}  & .852 & 0.843 & .855 & / &  / \\
            \bottomrule
        \end{tabular}
    }
    \caption{The results on the DUT-OMRON dataset of different SOD models and ExPert of metrics MAE, $F_\beta$, max E-measure and S-measure. The best results are in \textbf{bold}. }
    \label{More SOTA compare}
\end{table*}

\subsubsection{Qualitative Comparison}

\begin{figure}
    \centering
    \includegraphics[width=322pt]{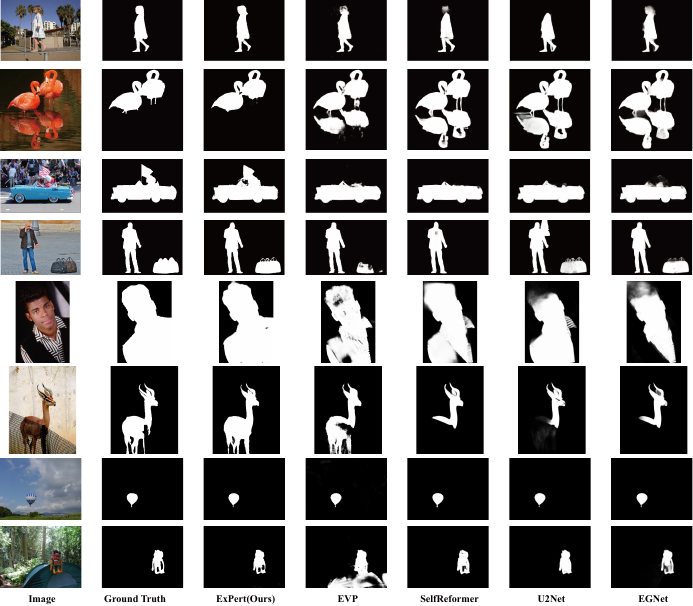}
    \caption{The qualitative results of ExPert and four SOTA models. From left to right are the images, the ground truths, ExPert's masks, EVP's masks, SelfReformer's masks, U2Net's masks and EGNet's masks. Better visual effect when zooming in.}
    \label{Qualitative}
\end{figure}

In Fig.~\textcolor{red}{\ref{Qualitative}}, we show the qualitative comparison between our model and SOTA models to give readers an intuitive comprehension. Compared to SOTA models, ExPert's masks are more accurate in details and can distinguish some ambiguous scenarios like reflection in water in the second row. The head of the girl is similar to the tree in the background in the first row while the hair and the cloth similar to the dark background are confusing in the fifth row. Four SOTA models can't differentiate these nuances but ExPert can handle the ambiguity. In the first row, ExPert's mask is even more accurate than the ground truth which contains all the hair. Moreover, the semantic information injection from BLIP aids ExPert in recognizing the relationship between objects in an image. For example, in the fourth row, the baggage on the ground is obviously related to the man. ExPert segments them out while some other SOTA models neglect the baggage or focus wrongly on the street lamp. Another example is the third row, the children in the car should be regarded as a whole with the car. Semantic information also assists ExPert in handling complex scenarios, such as shadow interference or color similarity, for instance, the case of discerning the body of the deer in the shadow. As for salient objects of small sizes, ExPert can well recognize them with clear details and less ambiguity as shown in the last two rows.

\subsection{Ablation study} \label{ablation}

\textbf{Tuning methods:}  To verify the effectiveness of E-adapter on fine-tuning models for salient object detection, the performances of different fine-tuning methods are evaluated on the ECSSD dataset. As shown in Table.~\ref{Ablation table}, full fine-tuning\footnote{The full fine-tuning method trains all the parameters of the backbone and decoder using the new datasets.} achieves the best performance but requires a large size of trained parameters. Training from scratch requires a large size of trained parameters with poor performances and head tuning\footnote{The head tuning method trains only the decoder while keeping the backbone frozen.} doesn't output satisfying results either. As for adapter tuning, the performances are close to the fully tuned version with much fewer trained parameters. 

\textbf{Single-scale and multi-scale backbone:} To compare the performance of the single-scale backbone and the multi-scale backbone, all fine-tuning methods are used to fine-tune MiT-B4 SegFormer and two of them are used to fine-tune ViT-B/16. The decoder of ViT is the same as~\cite{zheng2021rethinking_SETR}, more details can be found in our supplement file. In Table.~\ref{Ablation table}, training from scratch and adapter tuning of ViT are worse than the counterparts of SegFormer because multi-scale features can extract more detailed information in the images. Therefore, ExPert uses multi-scale pre-trained models as the encoder backbone for their finer details.

\begin{table*}
    \centering
    \resizebox*{1.0\columnwidth}{!}{
        \begin{tabular}{lrrrrr}
            \toprule
            Methods  & \quad MAE$\mathbf{\downarrow}$ & \quad max-FM$\mathbf{\uparrow}$ & \quad max-EM$\mathbf{\uparrow}$ & \hspace{0.3em} \quad\quad SM$\mathbf{\uparrow}$ & Trained Parameters \\
            \midrule
            Scratch-S   & 0.0784  & 0.8552 & 0.8758 & 0.8085 & 733.8MB \\
            Head tune-S   & 0.0740  & 0.8922 & 0.9292 & 0.8444 & 36.1MB \\
            Full tune-S   & \textbf{0.0282}  & \textbf{0.9501} & \textbf{0.9679} & \textbf{0.9320} & 733.8MB \\
            Adapter tune-S & 0.0354  & 0.9447 & 0.9626 & 0.9193 & 50.9MB \\
            Adapter tune-S-FFT  & 0.0383  & 0.9446 & 0.9612 & 0.9152 & 48.0MB \\
            From scratch-V   & 0.0312  & 0.9412 & 0.9648 & 0.9264 & 1093.7MB \\
            Adapter tune-V   & 0.0362  & 0.9300 & 0.9569 & 0.9163 & 130.3MB \\
            \bottomrule
        \end{tabular}
    }
    \caption{The results on ECSSD datasets of different fine-tune methods. Scratch refers to training from scratch. Head tune and full tune denote head tuning and full fine-tuning respectively. Adapter tune uses the block-level E-adapter to adapter tune pre-trained models. "-S" means the SegFormer backbone. "-V" means the ViT-B/16 backbone. "-FFT" means the adapter is of FFT level, otherwise the block-level. The metric in \textbf{bold} is the best.}
    \label{Ablation table}
\end{table*}

\textbf{Adapter level:}   In~\cite{chen2022adaptformer}, the adapter module is side connected inside the transformer block while in ExPert we use the block-level adapter, which is side connected between transformer blocks. In Table.~\ref{Ablation table}, the block-level adapter performs better than the FFT-level adapter which demonstrates that in the segmentation task the block-level adapters are more advantageous. We suppose that the self-attention layers are also important in image recognition and should also be covered by adapters.

\textbf{Prompt feature:}  To verify the effectiveness of E-injector and different prompt features, we denote the baseline model as the multi-scale encoder with E-adapter but without E-injector. In Table.~\ref{Ablation table of prompt feature}, we compared E-injector with five prompt features to the baseline. Injecting DINO features as prompt features performs better than the baseline, which shows that object-aware features from DINO can guide the encoder to focus more on salient objects. The ViT features injection is better than the DINO features injection, suggesting that pre-trained models' semantic information can further boost the performance of SOD. Additionally, although the multi-scale backbone can extract multi-scale salient features, its global layers are usually shallower than those in single-scale backbones. This weakness can be alleviated by the injection of ViT's features which go through more layers of the full-size scale.

Compared to the injection of ViT's features, the injection of BLIP's vision feature performs better. Since BLIP is trained with image-text pairs, the semantic recognition of BLIP is stronger than ViT which is trained with image-label pairs. This indicates that semantic information is critical for the model to detect salient objects. The BLIP+ injection version surpasses the BLIP injection version, verifying that our cross-attention interaction between BLIP's image features and caption embeddings successfully highlight the salient regions. The best performance comes from the combination of ViT's features and the BLIP+ features, which captures both the detailed large scale features and the rich semantic information.

\begin{table*}
    \centering
    \resizebox*{1.0\columnwidth}{!}{
        \begin{tabular}{lrrrrr}
            \toprule
            Methods  & \quad MAE$\mathbf{\downarrow}$ & \quad max-FM$\mathbf{\uparrow}$ & \quad max-EM$\mathbf{\uparrow}$ & \hspace{0.3em} \quad\quad SM$\mathbf{\uparrow}$ & Trained Parameters \\
            \midrule
            Baseline   & 0.0354   & 0.9447    & 0.9626   & 0.9193   & 50.9MB  \\  
            DINO Inject   & 0.0298  & 0.9473 & 0.9625 & 0.9330 & 60.0MB \\   
            ViT Inject   & 0.0272  & 0.9497 & 0.9654 & 0.9357 & 60.0MB \\    
            BLIP Inject   & 0.0249  & 0.9516 & 0.9681 & 0.9380 & 60.0MB \\    
            BLIP+ Inject   & 0.0231  & 0.9534 & 0.9689 & 0.9408 & 80.2MB \\     
            ViT \& BLIP+ Inject   & \textbf{0.0215}  & \textbf{0.9550} & \textbf{0.9707} & \textbf{0.9422} & 80.2MB \\   
            \bottomrule
        \end{tabular}
    }
    \caption{The results on ECSSD datasets of different prompt features for E-adapter. BLIP+ represents the interacted features after cross-attention of BLIP. The baseline only uses the E-adapter in~\ref{ExPert Adapter Chapter} without the E-injector.}
    \label{Ablation table of prompt feature}
\end{table*}

\section{Conclusion}

We introduce the \textbf{EX}ternal \textbf{P}rompt features \textbf{E}nhanced adapte\textbf{R} \textbf{T}uning (ExPert) model, designed to efficiently fine-tune pre-trained transformer models for salient object detection. E-adapter efficiently tailors pre-trained backbones to extract salient features, while E-injector integrates various external features as guiding prompts, enhancing the localization of salient objects. Additionally, to enhance the representation of fine details, ExPert incorporates ViT features into the backbone to complement shallow global layers. Furthermore, to capture the relationship between image content and salient elements, the image-text interaction features from BLIP are integrated into the encoder, enabling better differentiation of complex scenarios. Comprehensive experiments demonstrate the superior performance of our ExPert over both state-of-the-art CNN-based and transformer-based models across five validation datasets.

Looking ahead, further enhancement of ExPert may include exploration of additional prompt features. It is possible to inject other prompt features such as color or texture information into the backbone. Moreover, ExPert's paradigm could also be applied to other segmentation tasks, such as semantic segmentation and panoptic segmentation. We also find that ExPert's performance is influenced by the quality of the generated captions by BLIP. To make ExPert more robust, how to filter out captions of low quality is a challenge. We leave these possible directions for future research.

\clearpage


\title{External Prompt Features Enhanced Parameter-efficient Fine-tuning for Salient Object Detection \\ 
\emph{ Supplementary Materials}}
\titlerunning{Supplementary Materials}
%


\author{Wen Liang\inst{1} \and  
Peipei Ran\inst{2} \and  
Mengchao Bai\inst{2} \and 
Xiao Liu\inst{2} \and  
P. Bilha Githinji \inst{1} \and   
Wei Zhao \inst{2} \and  
Peiwu Qin\inst{1}\textsuperscript{\Letter}}  
\authorrunning{W. Liang et al.}
%
\institute{Tsinghua Shenzhen International Graduate School, Tsinghua University, Shenzhen, China \and
Central Media Technology Institute, Huawei, Shenzhen, China
}

\maketitle              

\textit{
This is the supplementary file for our ExPert model and additional experiment results. Contents are organized as follows:
\begin{itemize}
  \item[$\bullet$] More details of ExPert.~\ref{Supp ExPert details}
  \item[$\bullet$] More details of datasets and metrics.~\ref{Supp metric details}
  \item[$\bullet$] Additional ablation experiments.~\ref{Supp quantitative}
  \item[$\bullet$] Additional qualitative results.~\ref{Supp qualitative}
\end{itemize}
} 

\section{More details of ExPert}  \label{Supp ExPert details}

The input image size of our model is 384*384. If the original size is not 384*384, we resize the input images to the certain size. The 384*384 resolution is used for ViT-B/16~\cite{dosovitskiy2020image_ViT} and MiT-B4 version of SegFormer~\cite{xie2021segformer} for both training and inferring.

We also tried the single-scale pre-trained model ViT-B/16 as the backbone of ExPert. Since the ViT model is trained for classification tasks, the decoder should be changed to a matched one. A simple decoder is adopted for ViT which is drawn from SETR~\cite{zheng2021rethinking_SETR}. For $N(=12)$ ViT layers, $N/4$ layers are selected with an interval of 4 as the feature set ${Z_s}=\{f_2,f_5,f_8,f_{11}\}$ where $f_i$ is the vision feature of the $i_{th}$ layer. Let $\mathcal{U}()$ be the upsampling, $Cat()$ be the concatenate operation applied to all $f_i$ in $Z_s$ and $\mathcal{S}()$ be the Sigmoid Function. The Mask $M$ is generated as Eq.~\ref{vit mask cal} where $F_c$ is the concatenated feature. $Conv2()$ represents two convolution layers with an activation layer and $Conv()$ represents a single convolution layer.

\begin{align}
    F_c &=  Cat(\mathcal{U}(Conv2(f\in Z_s))) \\
    M &= \mathcal{S}(\mathcal{U}(Conv(F_c)))
\label{vit mask cal} 
\end{align}

\section{More details of datasets and metrics} \label{Supp metric details}

\textbf{Training dataset:} We use DUTS \cite{wang2017learning-DUTS} as our training dataset following the previous works. DUTS is the largest salient object detection benchmark that contains a training dataset and a test dataset, including respectively 10,553 images and 5,019 images. The training images are all from the ImageNet DET training or validation sets with slight differences in size. We use the test set of DUTS as our validation and evaluation dataset.

\textbf{Evaluation dataset:} We used ECSSD \cite{yan2013hierarchical-ECSSD}, DUT-OMRON \cite{yang2013saliency-DUT-OMRON}, HKU-IS \cite{li2015visual-HKU-IS} and PASCAL-S \cite{li2014secrets-PASCAL-S} as the evaluation datasets. The ECSSD dataset contains 1,000 real-world images across different domains with salient objects. DUT-OMRON is a challenging dataset containing 5,168 high quality images with complex backgrounds and various kinds of objects. The HKU-IS dataset has 4,447 well-annotated images characterized by multiple objects in diverse scenarios. PASCAL-S is a subset of the PASCAL VOC segmentation validation dataset, containing 850 images. Unless specified, we used ECSSD as our default evaluation dataset for ablation experiments.

\textbf{Evaluation metric:} Four metrics are adopted for our model evaluation. The first metric is the mean absolute error (MAE) as shown in Eq.~\textcolor{red}{\ref{eq_mae}}, which computes the average pixel difference between predicted maps and ground truth labels. A smaller value of MAE signifies better performance. 

\begin{align}
    MAE = \frac{1}{H \times W} \sum_{i=1}^{H}\sum_{j=1}^{W} \lvert pred(i,j) - gt(i,j) \rvert
\label{eq_mae} 
\end{align}

The second metric is F-measure $F_\beta$ \cite{achanta2009frequency_fmeasure} which is obtained by leveraging the precision and the recall as shown in Eq.~\textcolor{red}{\ref{eq_fmeasure}}. The precision and recall are computed using different thresholds from 0 to 255 between the predicted map and the ground truth. The $\beta^2$ is set to 0.3 following the former works. We use the maximum $F_\beta$ for the performance of each method. A larger value of $F_\beta$ signifies better performance. 

\begin{align}
    F_\beta = \frac{(1+\beta^2)Precision \times Recall}{\beta^2 \times Precision + Recall}
\label{eq_fmeasure} 
\end{align}

We also use the maximum E-measure \cite{fan2018enhanced_emeasure} and the S-measure \cite{fan2017structure_smeasure} as validation metrics. E-measure considers both pixel-level errors and image-level errors. S-measure is used to compare the structural similarity of the predicted map and the ground truth with region-aware and object-aware similarity. These two metrics are both better when their values are larger.

\section{Additional ablation experiments}  \label{Supp quantitative}

\subsection{Cross attention layers}

\begin{table*}
    \centering
    \resizebox*{1.0\columnwidth}{!}{
        \begin{tabular}{lrrrr}
            \toprule
            Methods  & \quad MAE$\mathbf{\downarrow}$ & \quad max-FM$\mathbf{\uparrow}$ & \quad max-EM$\mathbf{\uparrow}$ & \hspace{0.3em} \quad\quad SM$\mathbf{\uparrow}$ \\
            \midrule
            L=1   & \textbf{0.0231}  & \textbf{0.9534} & \textbf{0.9689} & \textbf{0.9408} \\
            L=2   & 0.0270  & 0.9465 & 0.9624 & 0.9349 \\
            L=4   & 0.0278  & 0.9475 & 0.9644 & 0.9332 \\
            L=8   & 0.0290  & 0.9503 & 0.9663 & 0.9315 \\
            \bottomrule
        \end{tabular}
    }
    \caption{The results on the ECSSD datasets of ExPert using E-adapter and E-injector with BLIP's interacted features. L refers to different numbers of cross attention layers. The best performance is in \textbf{bold}.}
    \label{cross attention ablation}
\end{table*}

As shown in Table.~\ref{cross attention ablation}, one cross attention is enough for the interaction and achieves the best performance. We assume that the model pre-trained by image-caption pairs does not need too much cross attention interaction to highlight the relevant regions in images of text. Since BLIP is trained with image-caption pairs, the alignment of image and text features is not of pixel scale but rather of patch scale. As auxiliary prompt features, too many cross attentions might lead to unexpected noises that harm the final performance.  Besides, fewer cross attention layers can further reduce the size of trained parameters. 

\subsection{Other transformer backbones}

ExPert is backbone-agnostic and can be extended to arbitrary transformer backbones. In the paper, we already showcased that ExPert works well with SegFormer and ViT which represent respectively the multi-scale and single-scale backbone. To further demonstrate ExPert's flexibility, we replace the SegFormer backbone with the Pyramid Vision Transformer (PVT) backbone\cite{wang2021pyramid_PVT}. The decoder head shares the same architecture as the multi-scale decoder in Fig. 2 of our paper. As shown in Table.~\ref{PVT compare}, E-adapter and E-injector boost the performance of the PVT backbone compared to full finetuning on SOD datasets with a much smaller size of training parameters which proves the flexibility of ExPert. However, the resolution of the PVT backbone is 224*224 which is smaller than the 384*384 of SegFormer. This might be the reason for the inferior performance of the PVT backbone compared to the SegFormer backbone. 

\begin{table*}
    \centering
    \resizebox*{1.0\columnwidth}{!}{
        \begin{tabular}{lrrrrr}
            \toprule
            Methods  & \quad MAE$\mathbf{\downarrow}$ & \quad max-FM$\mathbf{\uparrow}$ &\quad max-EM$\mathbf{\uparrow}$ & \quad SM$\mathbf{\uparrow}$ & \quad Size \\
            \midrule
            PVT-F   & 0.1689  & 0.7883 & 0.8396 & 0.6124 & 733.73MB \\
            PVT-A   & 0.0486  & 0.9213 & 0.9331 & 0.8842 & 50.85MB  \\
            PVT-E   & 0.0409  & 0.9299 & 0.9531 & 0.9026 & 89.26MB \\
            SegF-A   & 0.0354  & 0.9447 & 0.9626 & 0.9193 & 50.9MB \\
            SegF-E   & 0.0215  & 0.9550 & 0.9707 & 0.9422 & 80.2MB \\
            \bottomrule
        \end{tabular}
    }
    \caption{The results on the ECSSD dataset of the PVT backbone compared to SegFormer backbone. PVT represents the Pyramid Vision Transformer\cite{wang2021pyramid_PVT} backbone while SegF represents the SegFormer\cite{xie2021segformer} backbone. "-F" represents full finetuning, "-A" represents the adapter tuning with the frozen backbone and only E-adapter without E-injector. "-E" represents the ExPert framework with E-adapter and E-injector. The size column represents the size of trained parameters.}
    \label{PVT compare}
\end{table*}

\section{Additional qualitative results} \label{Supp qualitative}

\begin{figure}
    \centering
    \includegraphics[width=350pt]{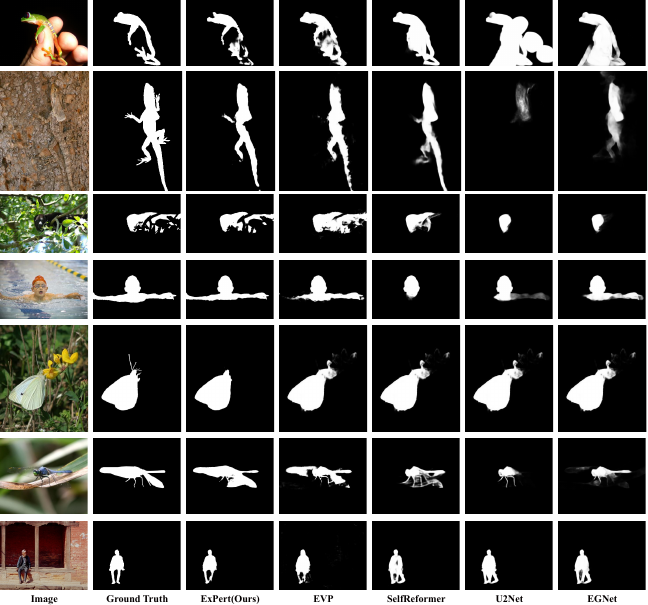}
    \caption{The qualitative results of ExPert and four SOTA models. From left to right are the images, the ground truths, ExPert's masks, EVP's masks, SelfReformer's masks, U2Net's masks and EGNet's masks. Better visual effect when zooming in.}
    \label{Supp Qualitative}
\end{figure}

Fig.~\ref{Supp Qualitative} shows some more examples of qualitative results between ExPert and four SOTA models. ExPert can differentiate the salient objects that are similar to or camouflaged in the background, for example, in the second row, the third row and the fourth row. Owing to the injection of semantic information, ExPert can handle the ambiguity between different objects in an image, for example, in the first row and the fifth row. What's more, the injection of ViT's feature makes up for the shallow layers of the full-size scale feature extraction stage of the multi-scale backbone. As a result, ExPert can discover some details hard to find. For example the nearly transparent wing of the insect in the sixth row. Finally, the injection of the complementary features and the semantic information can help ExPert differentiate the shadow distraction in the seventh row where other SOTA models struggle to discern the legs of the man. 

\clearpage

\bibliographystyle{splncs04}
\bibliography{mybibliography}

\end{document}